\documentclass[letterpaper, 10pt, conference]{ieeeconf}
\usepackage{subcaption}
\usepackage{multirow}
\usepackage{balance}
\usepackage{hyperref}
\usepackage{dsfont}
\usepackage{breqn}
\usepackage{makecell}
\usepackage{xcolor}
\usepackage{amsfonts}
\pdfoutput=1

\usepackage{booktabs}
\IEEEoverridecommandlockouts                              

\title{Design and Mechanics of Cable-Driven Rolling Diaphragm Transmission for High-Transparency Robotic Motion}

\author{Hoi Man Lam$^1$, W. Jared Walker$^1$, Lucas Jonasch$^1$, Dimitri Schreiber$^2$, \IEEEmembership{Student Member, IEEE}, \\and Michael C. Yip$^2$, \IEEEmembership{Senior Member, IEEE}
\thanks{$^1$Hoi Man Lam, W. Jared Walker, and Lucas Jonasch are with the Department of Mechanical and Aerospace Engineering, University of California San Diego, La Jolla, CA 92093 USA. {\tt\small hml024@ucsd.edu}, \tt\small \{wjaredw,lucasjonasch98\}@gmail.com}
\thanks{$^2$Dimitri Schreiber and Michael C. Yip are with the Department of Electrical and Computer Engineering, University of California San Diego, La Jolla, CA 92093 USA. {\tt\small \{dschreib, yip\}@ucsd.edu}}
}

\begin{document}

\maketitle

\begin{abstract}
Applications of rolling diaphragm transmissions for medical and teleoperated robotics are of great interest, due to the low friction of rolling diaphragms combined with the power density and stiffness of hydraulic transmissions.
However, the stiffness-enabling pressure preloads can form a tradeoff against bearing loading in some rolling diaphragm layouts, and transmission setup can be difficult.
Utilization of cable drives compliment the rolling diaphragm transmission's advantages, but maintaining cable tension is crucial for optimal and consistent performance.
In this paper, a coaxial opposed rolling diaphragm layout with cable drive and an electronic transmission control system are investigated, with a focus on system reliability and scalability.
Mechanical features are proposed which enable force balancing, decoupling of transmission pressure from bearing loads, and maintenance of cable tension.
Key considerations and procedures for automation of transmission setup, phasing, and operation are also presented.
We also present an analysis of system stiffness to identify key compliance contributors, and conduct experiments to validate prototype design performance.

\end{abstract}


\section{Introduction}
Transmissions are essential in systems where placing actuators at the joints is infeasible or dangerous. 
Examples of such systems include surgical robots or wearable robots, where high inertias on distal joints can pose dangers to the patient or user. 
Another example is MR-compatible surgical robots, where actuators are incompatible and can disturb the magnetic fields used for imaging.

Hydraulic transmissions are a viable solution to these problems, offering high power density, routing flexibility, and high stiffness.
Recent usage of rolling diaphragms in hydraulic transmissions provide high force transparency and bandwidth \cite{bolignariDesignExperimentalCharacterization2020}, which are important advantages in medical robotic applications.

However, current research applications of rolling diaphragm transmissions usually introduce a tradeoff between beneficial transmission pressures versus detrimental frictional forces and torque loading.
Furthermore, the fluid transmission's complex setup procedure is a detriment to the adoption of rolling diaphragm transmissions beyond research.

\begin{figure}[t!]
    \vspace{2mm}
    \centering
    \includegraphics[width=0.96\columnwidth, trim={0 0 0 0 mm}, clip=true]{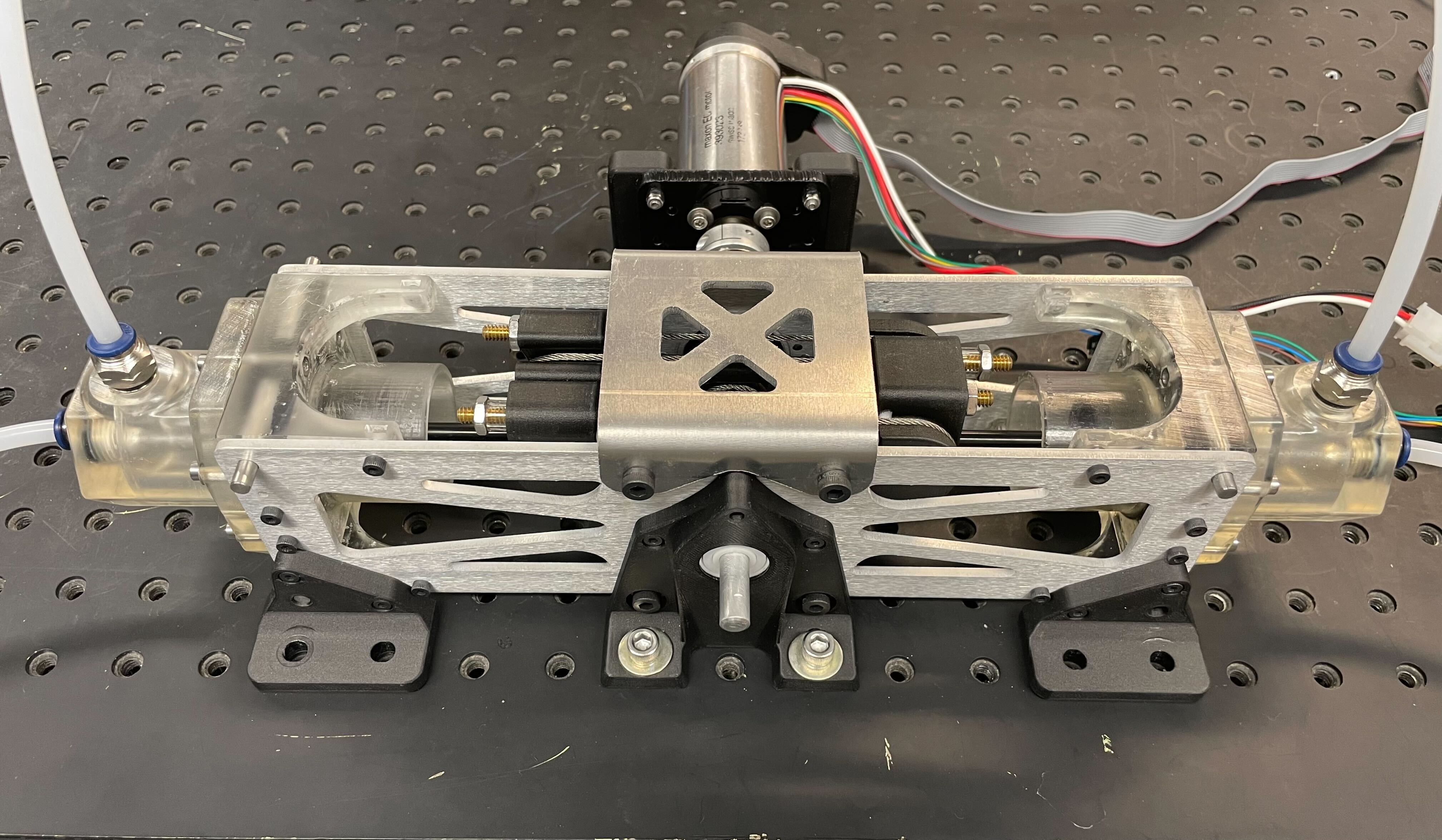}
    \caption{Image of one prototype transmission unit with a motor attached to the input/output shaft. Two of these units connected through fluid lines form a full transmission.}
    \label{fig:cover}
\end{figure}

\subsection{Related Works}
Rolling diaphragms have been applied to many applications, such as in medical or teleoperated robotics. 
In medical applications, rolling diaphragm transmissions have found great use in MR-safe robots, which make surgery during live scans possible by allowing the MR-incompatible actuator to power the robot from outside the MR field \cite{dongHighPerformanceContinuousHydraulic2019}, \cite{dongRoboticCatheterSystem2017}, \cite{guoCompactDesignHydraulic2018},
\cite{frishmanExtendingReachMRI2021}.
The usage of rolling diaphragm transmissions also provides smooth force transmission in a compact package, which is important for patient comfort, and accommodation of a variety of patient sizes.

The hydrostatic rolling diaphragm transmission itself was investigated in various works, where some works modelled the transmission as a second-order spring model
\cite{dongHighPerformanceContinuousHydraulic2019},
\cite{gruebeleLongStrokeRollingDiaphragm2019}, 
\cite{denisLowLevelForceControlMRHydrostatic2021}.
John Peter Whitney proposed a N+1 hybrid hydrostatic transmission setup, utilizing N hydraulic lines and 1 common pneumatic preload line, which minimizes transmission complexity without hampering performance \cite{whitneyHybridHydrostaticTransmission2016}, \cite{whitneyLowfrictionPassiveFluid2014},
\cite{mendozaTestbedHapticMagnetic2019}. 

Whitney’s designs featured a belt-driven design for rotary actuation, and a linkage design for directly actuating a finger, to convert the transmission’s translational motion into the desired mechanical work. 
In the linkage hand design, the load is attached to the diaphragm piston through the air-side diaphragm, introducing a dynamic seal, which can be tolerance-intensive and introduces constant pressure leakage.

In a hydrostatic transmission used for wearable robotic limbs, a ball-screw design is used on the actuation side, whilst the manipulator side uses a rolling diaphragm \cite{veronneauMultifunctionalRemotelyActuated2020}, \cite{9128040}. 
A cable drive provides reduced friction and backlash to accentuate the rolling diaphragm transmission’s transparency and backdrivability, while the floating-cylinder layout makes use of the self-aligning features of the rolling diaphragm.

\subsection{Contributions}
In this work, we investigate an alternative coaxial rolling diaphragm transmission design, which retains proven features such as the N+1 hybrid transmission layout, and a cable drive for its ability to minimize backlash and friction forces. Building on top of those elements, we introduce the following:
\begin{enumerate}
    \item A translating inner core inline with the rolling diaphragm, coupled with a force-balanced angled cable drive design, which decouples transmission pressure from cable tension and friction, reduces bearing load thanks to cable preload tension balancing, and provides a constant mechanical advantage over a range of motion beyond a full rotation.
    \item A method for individual transmission component stiffness analysis to predict full system stiffness and key compliance contributors.
    \item An electronic fluid transmission control system for automated pressure and phasing regulation and control.
\end{enumerate}




\section{Method}

\subsection{Overview}
\begin{figure}[t!]
    \vspace{3mm}
    \centering
    \begin{subfigure}[b]{1\columnwidth}
        \includegraphics[width=0.96\columnwidth, trim={0 0 0 0 mm},clip=true]{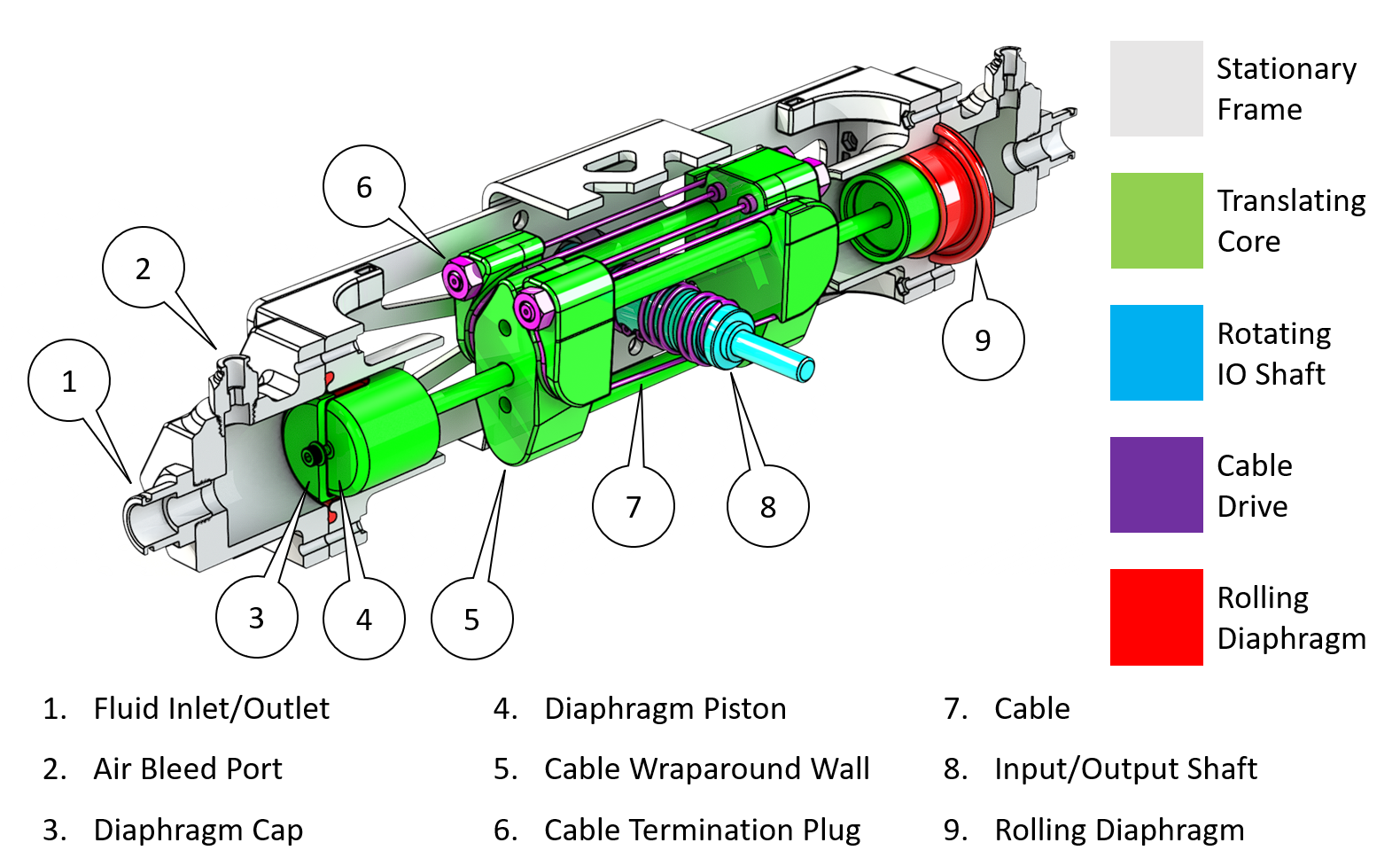}
        \caption{Section view of one actuator unit.}
        \label{fig:sectionView}
    \end{subfigure}
    \begin{subfigure}[b]{1\columnwidth}
        \centering
        \includegraphics[width=0.96\columnwidth, trim={0 0 0 0 mm}, clip=true]{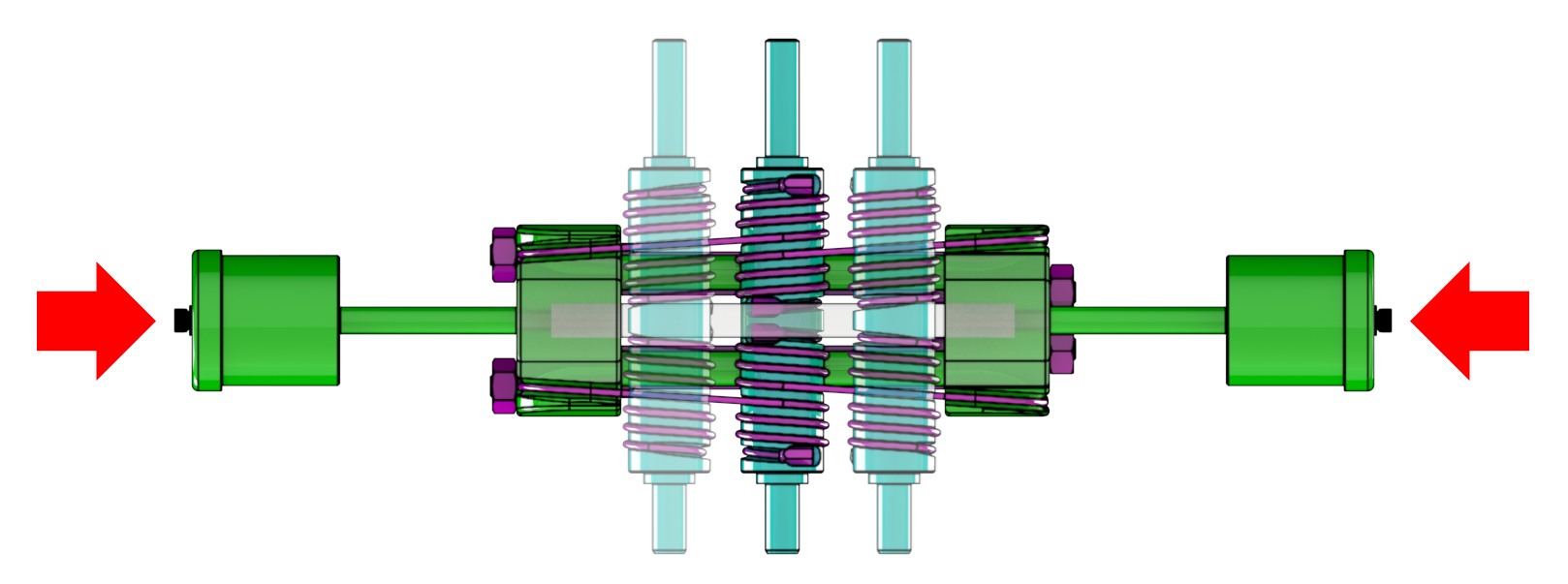}
        \caption{Translating capstan pillar relative to the core.}
        \label{fig:fleet angle}
    \end{subfigure}
    \caption{Mechanical design of one actuator unit. (a) shows the section view. Fluid pressure difference between the two rolling diaphragms (red) actuate the translating core (green), which drives the input/output shaft (cyan) through the cable drive (purple). The coaxial rolling diaphragms are fully sealed, and the translating core maintains cable tension even when the system is depressurized. (b) shows how the capstan pillar rolls relative to the core. 
    Fixed to the housing via bearings, the pillar no longer translates, and is limited to rotation as the core translates.
    Cable drive fleet angle is kept constant across the full range of motion through the geometry of the capstan pillar and the cable wrap-around walls, maintaining constant cable tension for controllability. Fluid line pressure preload forces are coaxially balanced through the translating core.}
\end{figure}

The transmission consists of two identical paired units, where each unit uses two rolling diaphragms to interface with the two air or water transmission lines. 
%
In each transmission unit (Fig. \ref{fig:sectionView}), a cable drive converts the fluid-driven linear rolling diaphragm motion to rotary input-output motions for robotic joints. 
A linearly translating core actuates that cable drive, while isolating the transmission pressure forces from the cable tension and central pillar bearings. 
Lastly, a stationary frame provides structural support and alignment between rolling diaphragms and translating core.

\subsection{Cable Drive and Translating Core}
A cable drive was chosen due to its inherent advantages of low friction, low backlash, and high stiffness \cite{BASER2010815}, \cite{12057}. 
In the prototype transmission, a 1/16” diameter 302 stainless steel cable with 7x19 construction was chosen for its high flexibility, and sufficient breaking strength of $2100N$ to match the maximum supportable load of $600N$. 
The maximum supportable force is a function of maximum rated diaphragm pressure $P_{max} = 1.7MPa$, and diaphragm piston radius $r_{piston}=15mm$ in Eqn. \ref{eqn:maxForce}. 
The corresponding maximum supportable torque is $6Nm$ for a capstan radius of $r_{capstan}=10mm$.
\begin{equation}\label{eqn:maxForce}
    F_{max} = \frac{P_{max}}{2}(\pi r_{piston}^{2})
\end{equation}
Due to the limited travel of the rolling diaphragm, there is a direct tradeoff between the diameter of a flat cable pulley and the angular range of motion. 
However, wrapping the cable helically up a capstan pillar allows for increased capstan effect and more than one rotation of motion despite limited diaphragm travel. 


Within the cable drive, the cables are run at an angle enforced by wall and capstan geometry throughout the full range of motion (Fig. \ref{fig:fleet angle}). 
The fleet angle is kept constant by terminating the cables with the same angle as the helix pitch. 
Constant fleet angle is necessary to maintain constant cable tension for a linear performance, which is useful in direct teleoperation applications.

\begin{figure}[t!]
    \vspace{2mm}
    \centering
    \includegraphics[width=0.7\columnwidth, trim={0 0 0 0 mm}, clip=true]{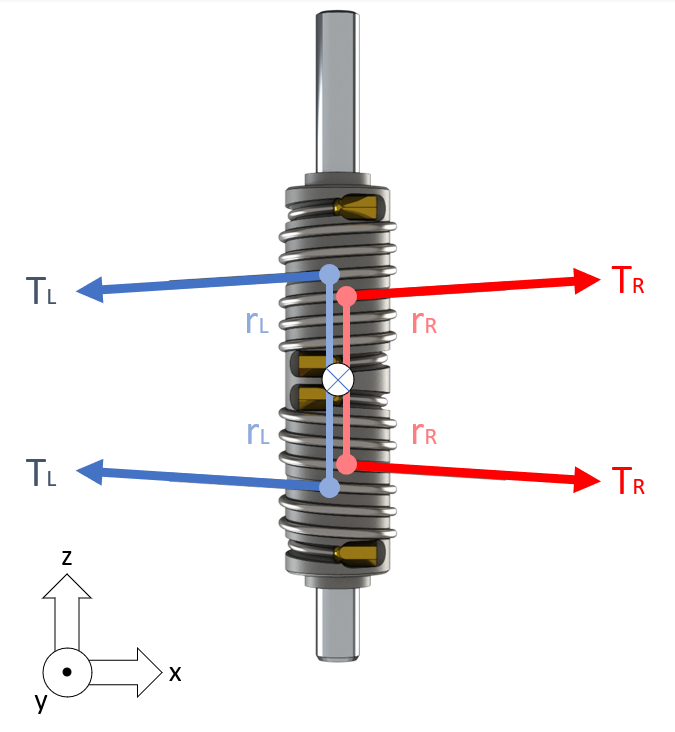}
    \caption{Cable tension forces are symmetrically balanced, thereby eliminating moment bearing loads in the y-axis (out of page) and decoupling cable tension from axial bearing loads in x/y directions.}
    \label{fig:balanced capstan}
\end{figure}

The cable drive’s forces are balanced by the symmetrical placement of cable angle and departure locations, which decouples cable preload tension from bearing loading. 
The free body diagram moment balance of Fig. \ref{fig:balanced capstan} sums to 0, indicating full cancellation:
\begin{equation}\label{eqn:tensionBalance}
    \Sigma M_{y} = T_{L}r_{L} - T_{L}r_{L} + T_{R}r_{R} - T_{R}r_{R} = 0
\end{equation}
Where $T_{L}$ and $r_{L}$ are the cable preload tension force and moment arm on the left side of the capstan, and $T_{R}$ and $r_{R}$ on the right side.
High cable preload tension is preferred to improve axial stiffness, which decreases with load range to preload ratio \cite{HOBBS199631}.

The translating core structure fixes the cable terminations and absorbs the transmission pressure between the two rolling diaphragms. 
One drawback to cable drives can be the difficulty in tuning and maintaining cable tension, which is a prerequisite for consistent performance. 
By fixing the cable terminations within the rigid core structure, the cables can be preloaded at high tension, allowing for low slack and high stiffness \cite{BASER2010815}. 
Cable tension is held even when the transmission is depressurized, which is advantageous in medical settings, where surgeons might not have the expertise or tools to troubleshoot and repair cable slack issues.
The prototype translating core comprises of solid 3D printed termination walls in Markforged Onyx filament and carbon fiber rods.


\subsection{Coaxial Enclosed Rolling Diaphragm Layout}
Interacting with the fluid transmission lines, rolling diaphragms interface with coaxial pistons on either side of the translating core. 
Rolling diaphragms are flexible seals that extend and retract via a rolling action, eliminating the sliding friction that typical piston O-Ring seals have \cite{marshBellofram}.  
The rolling diaphragms used in the prototype are DM3-35-35 rolling diaphragms manufactured by IER Fujikura, with stroke of 46mm, diameter of 35mm, and maximum pressure rating of $1.7MPa$.

Both fluid transmission lines have a pressure preload, which fills out the rolling diaphragm's convolution to prevent jamming, enables high fluid stiffness, and determines transmission load capacity.
This transmission preload pressure has a much higher magnitude of force relative to the magnitude of input/output force transferred. 

In the proposed transmission, the two rolling diaphragms are aligned in a coaxial opposed layout, such that preload pressures are balanced against one another (Fig. \ref{fig:fleet angle}). 
These preload pressure forces compress the translating core, but are isolated from bearings by the cable drive, decoupling transmission pressure from bearing friction.
The cable drive is situated between the diaphragm pistons, keeping both fluid chambers fully enclosed to avoid pressure leakage.

\subsection{Hydrostatic Transmission}

\begin{figure}[t!]
    \vspace{2mm}
    \centering
    \includegraphics[width=.99\columnwidth, trim={0 0 0 0 mm}, clip=true]{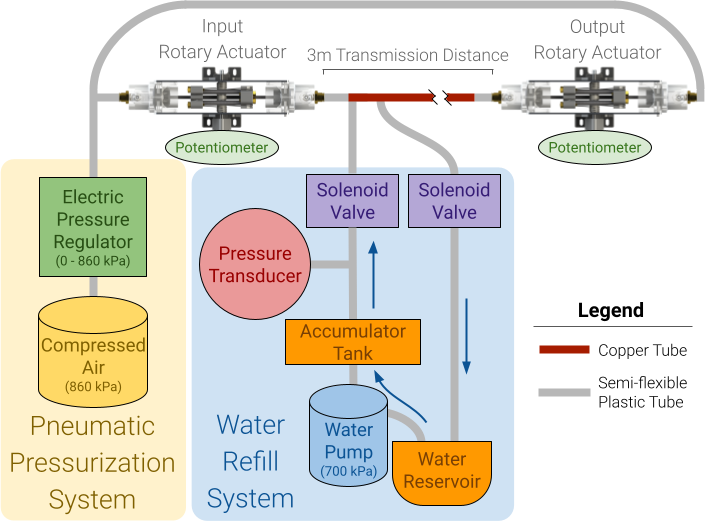}
    \caption{Transmission system setup. Stiffness is maximized along the hydraulic transmission distance by using a copper tube, while small sections of flexible plastic tube at each end allow for some flexibility in actuator placement. The incoming water line is maintained at $700 kPa$ by a water pump. The outgoing line releases into a depressurized reservoir which feeds into the pump. The preload pressure is controlled through a proportional electric pressure regulator, which allows for precise pressure control anywhere from 0 to $860 kPa$. The design is intended for an N+1 configuration (introduced in \cite{whitneyHybridHydrostaticTransmission2016}), such that multiple transmissions are preloaded by one single pneumatic line, simplifying scalability.}
    \label{fig:TransmissionOverview}
\end{figure}

Two identical actuators connected by a hydraulic line form one transmission system, detailed in Fig. \ref{fig:TransmissionOverview}. 
The hydraulic line acts as an incompressible link between the actuator, while an opposing pneumatic line provides a preload pressure on the water line.
High preload pressures help dissolve remaining air into the water to achieve a stiff and responsive system.

The volume of water in the hydraulic line determines the phase offset between the input and output shafts. 
Manual alignment is time consuming and difficult, and misaligned actuators may hit their endstops prematurely resulting in reduced range of motion and control. 
Automation of this phasing process removes a large hurdle for system adoption, and effectively negates long-term issues such as minuscule leaks and settling.
It also reduces misalignment factors like tube and cable flex by allowing phasing at a high pressure, close to the standard operating preload pressure.

\subsection{Predictive Phasing}
The microcontroller phases the actuators efficiently through a proportional controller, calculating the adjustment that is needed to rotationally align the actuators. The flow rate $Q$ through a solenoid valve is given by:

\begin{equation}\label{eqn:Flowrate}
    Q = K_v \sqrt{\Delta P}
\end{equation}

where \(\Delta P\) is the pressure drop and \(K_v\) the flow factor of the valve. 
The relationship between water volume and phase offset was determined empirically, resulting in the following relationship:

\begin{equation}\label{eqn:AngleWater}
    V_W = \frac{|\Delta \phi |}{9.594}
\end{equation}


where \(V_W\) is water volume in mL and \(\Delta \phi\) is the phase offset in degrees. 
By combining Eq. \ref{eqn:Flowrate} and \ref{eqn:AngleWater}, the time the solenoid valve should be opened is found. 
The sign of the phase offset \(\Delta \phi\) determines if the intake or outlet valve should be used.

\begin{equation}\label{eqn:TimeValveOpen}
    t = \frac{V_W}{Q} = \frac{|\Delta \phi |}{9.594K_v \sqrt{\Delta P}}
\end{equation}

\begin{figure}[t!]
    \vspace{3mm}
    \centering
    \includegraphics[width=0.7\columnwidth, trim={0 0 0 0 mm}, clip=true]{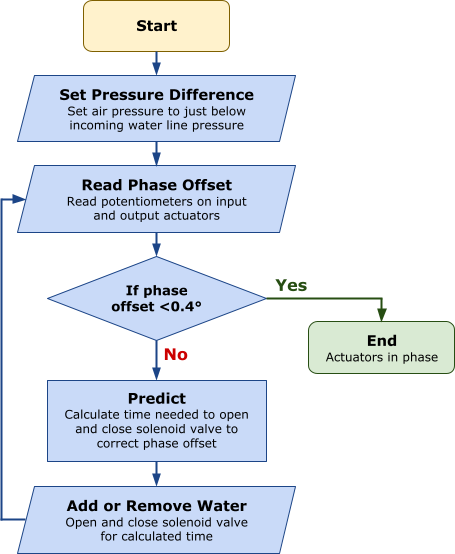}
    \caption{Automatic phasing algorithm. \(\Delta P\) of $15 kPa$, along with potentiometer accuracy and solenoid valve response time, resulted in a 0.4° acceptable maximum phase offset for our setup.}
    \label{fig:AutoPhaseLoop}
\end{figure}

Fig. \ref{fig:AutoPhaseLoop} shows the actuator phasing algorithm. 
The alignment resolution is limited by the \(\Delta P\) from the water refill system to the transmission line. 
To achieve finer angle adjustments, the transmission pressure is brought to just below the water injection pressure to decrease \(\Delta P\). 
This increases \(\Delta P\) for water ejection, but it is inconsequential since any overshoot can be corrected by water injection.

\subsection{Automated Operation}
\begin{figure}[t!]
    \vspace{3mm}
    \centering
    \includegraphics[width=0.96\columnwidth, trim={0 0 0 0 mm}, clip=true]{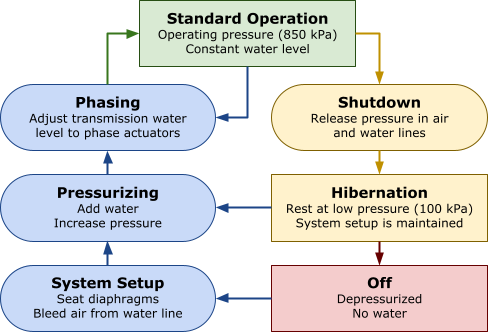}
    \caption{Operating procedure for the transmission system. Ovals represent transitory steps that lead to steps intended to be used for extended periods of time, represented by rectangles.}
    \label{fig:OperationLoop}
\end{figure}

The system operating procedure is outlined in Fig. \ref{fig:OperationLoop}. 
Users operate the system through a text-based interface with a microcontroller. Plain language instructions and commands allow even a non-technical user to set up and adjust a transmission.
An air bleed mode removes air bubbles that entered the water line during assembly. When not in operation, the system is stored in a hibernation state rather than completely removing all water and air pressure. Maintaining the system at a low pressure (for example, $100 kPa$) removes the need to bleed air from the water lines or reseat the rolling diaphragms. Completely depressurizing the actuator is used to disassemble or move the transmission setup.

\section{Theoretical Transmission Stiffness}

\begin{figure}[t!]
    \vspace{2mm}
    \centering
    \includegraphics[width=0.96\columnwidth, trim={0 0 0 0 mm}, clip=true]{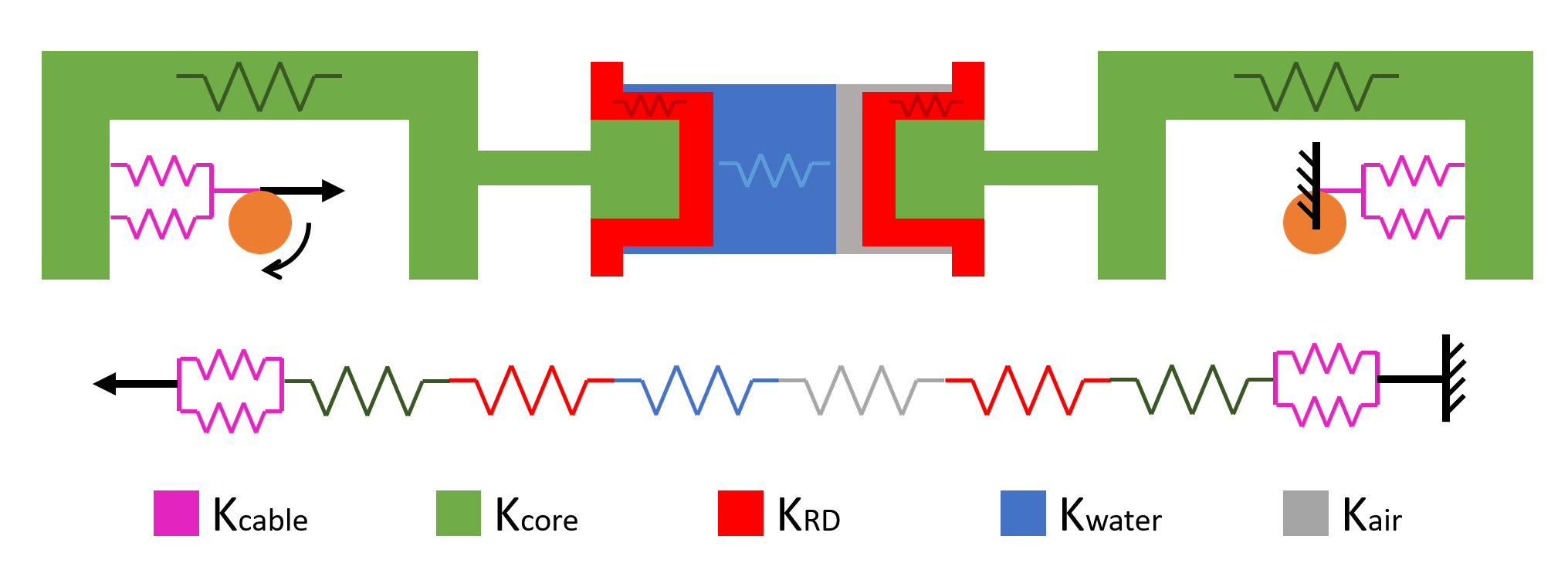}
    \caption{Transmission stiffness system model, assuming locked output shaft and torqued input shaft applying a force into the system. The model assumes some proportion of undissolved air left in the system, which acts as an additional spring in series in the fluid transmission.}
    \label{fig:stiffnessSystem}
\end{figure}

To predict the transmission stiffness and identify the main compliance contributors in the system, a simple spring system, as illustrated in Fig. \ref{fig:stiffnessSystem}, was used to model the transmission. 
Fluid stiffness and cable stiffness are determined analytically, while more complex components such as the translating core and diaphragm stiffnesses are determined respectively using FEA and empirical methods. 
The system assumes that the change in force is applied by the input side capstan pillar, while the output side capstan pillar is locked.

The fluid transmission stiffness is separated into a water stiffness (\begin{math}K_{water}\end{math}) and air stiffness (\begin{math}K_{air}\end{math}), assuming there is some proportion of undissolved air \begin{math}p_{air}\end{math} in the system. 
These fluid stiffnesses were found using Eq. \ref{eqn:KFluid} and the variables in Table \ref{tab:variables}.

\begin{equation}K_{fluid}=[p (2\frac{L_{cyl}}{A_{cyl}E_{fluid}} + \frac{L_{hose}}{A_{hose}E_{fluid}})]^{-1} \label{eqn:KFluid}
\end{equation}

\begin{table}[h]
    \vspace{2mm}
    \centering
    \caption{Theoretical fluid line stiffness variables}
    \def\arraystretch{0.8}
    \begin{tabular}{lll}
    \toprule

    \textbf{Variable}  & \textbf{Value} & \textbf{Description} \\ 
    \midrule
        \begin{math}E_{water}\end{math}   & 2.20 [\begin{math}GPa\end{math}]      & Bulk modulus of water \\
        \begin{math}E_{air}\end{math}     & 1.42 [\begin{math}GPa\end{math}]      & Bulk modulus of air \\
        \begin{math}A_{cyl}\end{math}     & 9.62E-4 [\begin{math}m^{2}\end{math}] & Cross section of diaphragm cylinder \\
        \begin{math}A_{hose}\end{math}    & 3.17E-5 [\begin{math}m^{2}\end{math}] & Cross section of hose \\
        \begin{math}L_{cyl}\end{math}     & 3.80E-2 [\begin{math}m\end{math}]     & Length of diaphragm cylinder \\
        \begin{math}L_{hose}\end{math}    & 4.26E-2 [\begin{math}m\end{math}]     & Length of hose \\
        \begin{math}p_{water}\end{math}   & 0.99 [\%]   & Proportion of water in transmission \\
        \begin{math}p_{air}\end{math}     & 0.01 [\%]   & Proportion of air in transmission \\
    \midrule
        \begin{math}K_{water}\end{math}   & 1.58E6 [\begin{math}N/m\end{math}]  & Est. K of fluid line water\\
        \begin{math}K_{air}\end{math}     & 3.99E3 [\begin{math}N/m\end{math}]  & Est. K of fluid line undissolved air\\
    \bottomrule
    \end{tabular}
    \label{tab:variables}
\end{table}

Cable stiffness (\begin{math}K_{cable}\end{math}) was calculated with Eq. \ref{eqn:cableStiff} 
Where \begin{math}A_{cable}\end{math} is the cable's cross-section, \begin{math}L_{cable}\end{math} is the cable length between the capstan and the wrap-around wall, and \begin{math}E_{cable}\end{math} is $200 GPa$ for 304 stainless steel.
\begin{equation}K_{cable}=\frac{E_{cable}A_{cable}}{L_{cable}} \label{eqn:cableStiff} 
\end{equation}

\begin{figure}[h]
    \centering
    \includegraphics[width=0.96\columnwidth, trim={0 0 0 0 mm}, clip=true]{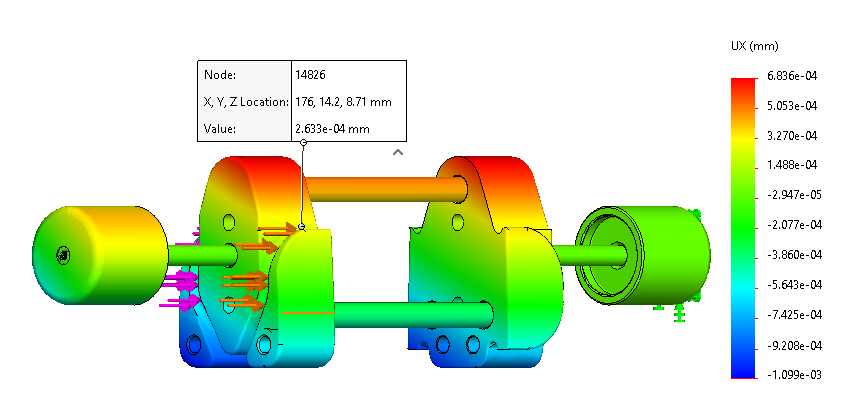}
    \caption{FEA estimated cable termination deflection from application of $1 N$ from cables.}
    \label{fig:coreFEA}
\end{figure}

For both rolling diaphragm stiffness (\begin{math}K_{RD}\end{math}) and translating core stiffness (\begin{math}K_{core}\end{math}), the stiffness was estimated via \begin{math}K = \frac{\Delta F}{\Delta x}\end{math}, where \begin{math}\Delta F\end{math} and \begin{math}\Delta x\end{math} are the change in force and deflection respectively.
For \begin{math}K_{RD}\end{math}, the rolling diaphragm stretch under force application was measured on a material test system. 
And for \begin{math}K_{core}\end{math}, the deflection of the cable termination point under a unit applied force was estimated via FEA (Fig. \ref{fig:coreFEA}). 

The resultant transmission stiffness estimate is a combination of the individual component stiffness estimates: 

\begin{equation} \label{eqn:Ktot}
K_{tot}=(\frac{1}{K_{cable}} + \frac{1}{2K_{core}} + \frac{2}{K_{RD}} + \frac{1}{K_{water}} + \frac{1}{K_{air}})^{-1}
\end{equation}    

\begin{figure}[h]
    \vspace{2mm}
    \centering
    \includegraphics[width=0.96\columnwidth, trim={0 0 0 0 mm}, clip=true]{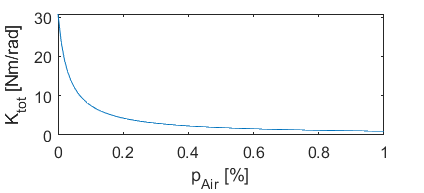}
    \caption{Theoretical transmission stiffness as a function of undissolved air \% remaining in fluid line.}
    \label{fig:airVstiffness}
\end{figure}

Where the individual estimates and overall stiffness estimate are featured in Table \ref{tab:stiffnesses}. 
Of the individual stiffnesses, the rolling diaphragm and undissolved air contribute the most to overall compliance.
The equivalent estimated angular stiffness over a $20 mm$ diameter capstan pillar is $23.54 Nm/rad$, using \begin{math} \label{eqn:KrotConvert} K_{rot}=K_{lin} r^{2}\end{math}.  
However, the stiffness value can vary greatly with \begin{math}p_{air}\end{math}, as shown in Fig. \ref{fig:airVstiffness}.

\begin{table}[h]
    \vspace{2mm}
    \centering
    \caption{Theoretical component stiffness values, and resultant total transmission stiffness assuming \begin{math}p_{air}\end{math} of 0.01 \%.}
    \def\arraystretch{0.75}
    \begin{tabular}{m{2.2cm} l m{2cm} m{1.8cm}}
    \toprule
    \textbf{Source} &\textbf{Variable}  & \textbf{Stiffness [\begin{math}N/m\end{math}]} & \textbf{\% Compliance} \\ 
    \midrule
        Water in fluid line             & \begin{math}K_{water}\end{math}    & 1.54E6  & 15.2\% \\
        Undissolved air in fluid line   &\begin{math}K_{air}\end{math}       & 9.97E5  & 23.6\% \\
        Single cable                    &\begin{math}K_{cable}\end{math}     & 8.98E6  & 2.60\% \\
        Translating core                &\begin{math}K_{core}\end{math}      & 3.80E6  & 12.4\% \\
        Rolling diaphragm               &\begin{math}K_{RD}\end{math}        & 1.02E6  & 46.1\% \\
    \midrule
        Full Transmission               & \begin{math}K_{total}\end{math}     & 2.35E5 & \\

    \bottomrule
    \end{tabular}
    \label{tab:stiffnesses}
\end{table}

\section{Experiments and Results}
\subsection{Experimental Setup
}

To characterize the system, experiments were conducted to fit a second-order spring model, and to understand the hysteresis and friction of the system. 
On the input side, the input shaft is actuated either by hand or by a motor (Maxon EC393023). 
Torque inputs are measured through a torque sensor (Futek TRS600) in the hand-actuated case, and estimated via the motor current draw in the motor actuated case. 
On the output side, a second torque sensor at the output shaft measures the torque output. 
Encoders (US Digital E5-2000) are fitted to both shafts to measure the angular deflection both into and out of the transmission. 

The effects of hose length and diameter were minimized in this experiment by using a minimum hose length of $42.6 mm$, to focus on the characteristics of the rolling diaphragm and cable drive.
The system is bled of air bubbles until no more air can be visually seen in the system.
The stiffness of shaft couples and torque sensors facilitating measurement are stiff enough to be disregarded in the stiffness estimation. 

\begin{figure}[h]
    \vspace{2mm}
    \centering
    \includegraphics[width=0.96\columnwidth, trim={0 0 0 0 mm}, clip=true]{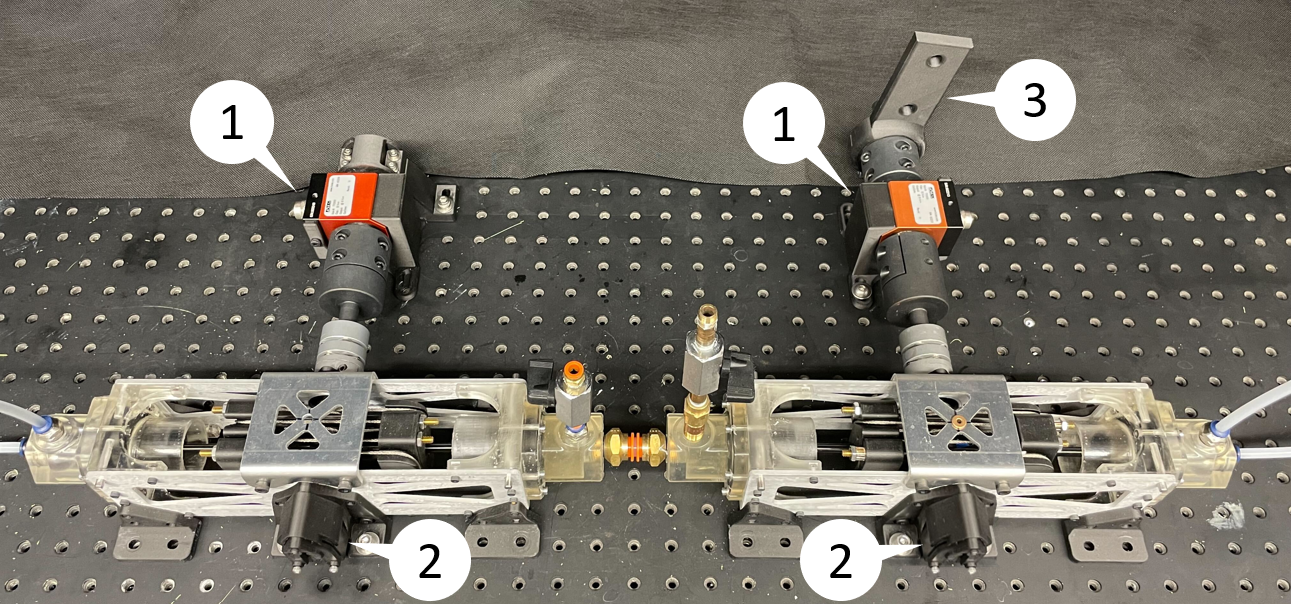}
    \caption{Experimental setup for hand-driven configuration. Shaft torques and positions are measured via torque sensors (1) and encoders (2). The input shaft is driven by a handle (3), or a motor can be substituted for motor-driven inputs.}
    \label{fig:ExperimentSetup}
\end{figure}

To estimate the second-order spring model, the output shaft is clamped, and the input shaft is driven with a consecutive torque step signal.
A \textit{tfest} is performed with measured torque and position as input and output (Eq. \ref{eqn:transferFunc}), starting from an initial model based on parameters from prior theoretical calculations (Table \ref{tab:modelFitResults}).
\begin{equation} \label{eqn:transferFunc}
H(s) = \frac{\theta(s)}{\tau(s)} = \frac{1}{Js^{2} + Bs + K}
\end{equation}

\begin{figure}[h]
         \centering
        \includegraphics[width=0.95\linewidth, trim={6mm 1mm 20mm 0mm}, clip=true]{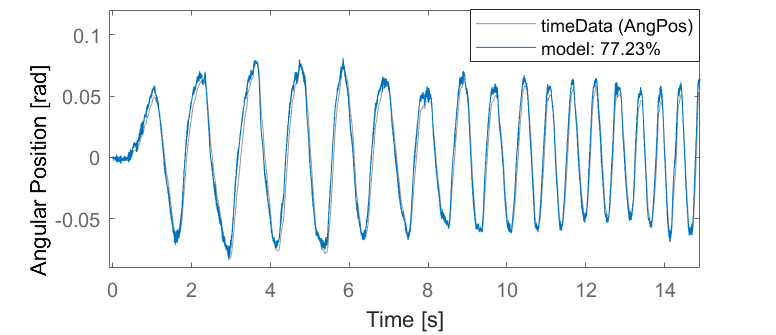}
         \label{fig:handVerify}
    \caption{Verification of fitted model against the hand-driven dataset, where the model predicted position from measured input torque is compared against the measured position, exhibiting accurate tracking and good model fit.}
    \label{fig:stepFitVerify}
    \vspace{-3mm}
\end{figure}

The resultant fit (Table \ref{tab:modelFitResults}) estimates a stiffness of $18.71 Nm/rad$, damping of $2.1 mNm/s$, and inertia of $0.52 \mu kg m^{2}$.
This lines up with the theoretical stiffness value corresponding to 0.02\% undissolved air in the fluid line (see Fig. \ref{fig:airVstiffness}), which is reasonable given that visually no air bubbles were left in the transmission, but some imperceptible undissolved air might still remain.
Additionally, hand-actuated datasets were collected for model validation, measuring input torque applied and input shaft deflection.
The model predicted position tracks the measured position closely in Fig. \ref{fig:stepFitVerify}, indicating model accuracy. 

\begin{table}[h]
    \centering
    \caption{Fit results of mass-spring-damper model on step dataset}
    \def\arraystretch{1}
    \begin{tabular}{p{.4\linewidth}p{.4\linewidth}}
        \hline
            Initial Model Coeffs                         & Step Fitted Model Coeffs\\ 
        \hline
            J = 6.54E-5 \begin{math}[kg m^{2}]\end{math} & J = 5.20E-5 \begin{math}[kg m^{2}]\end{math} \\
            B = 0.005  \begin{math}[Nm/s]\end{math}      & B = 0.0021  \begin{math}[Nm/s]\end{math}\\
            K = 23.54  \begin{math}[Nm/rad]\end{math}     & K = 18.71  \begin{math}[Nm/rad]\end{math}\\
        \hline
    \end{tabular}
    \label{tab:modelFitResults}
    \vspace{-3mm}
\end{table}

\begin{figure}[h]
    \centering
    \includegraphics[width=0.95\linewidth, trim={6mm 0 50 5mm}, clip=true]{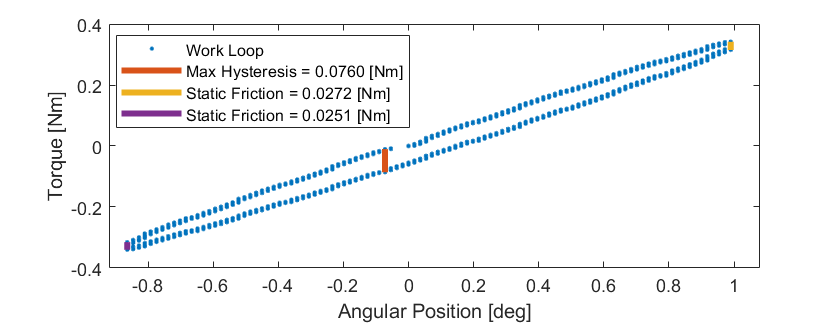}
    \caption{Hysteresis plot of one torque sin wave input starting from rest, maximum hysteresis measured is $0.0760Nm$ (1.27\% of $6Nm$ full torque range), and maximum static friction measured is $0.0272Nm$ (0.45\% of $6Nm$ full torque range).}
    \label{fig:Hysteresis}
\end{figure}

From a hysteresis plot of one torque sin wave motion from rest (Fig. \ref{fig:Hysteresis}), there is an approximate maximum hysteresis value of $0.076Nm$, corresponding to 1.25\% of the full $6Nm$ torque range. 
This hysteresis is likely caused by hose flexibility or air line pressure regulation inaccuracy, which affects shaft phasing, absorbs input energy, and has directional behaviour under positive and negative pressures.
The static friction, identified by the change in torque without change in angular position at either end of the hysteresis curve, is $0.025Nm$, corresponding to 0.45\% of the full torque range. 
Though the rolling diaphragm has little static friction, some other elements such as the cable drive and bearings still contribute to static friction.

\begin{figure}[t!]
    \vspace{2mm}
    \centering
    \includegraphics[width=0.95\linewidth, trim={20 0 50 20 mm}, clip=true]{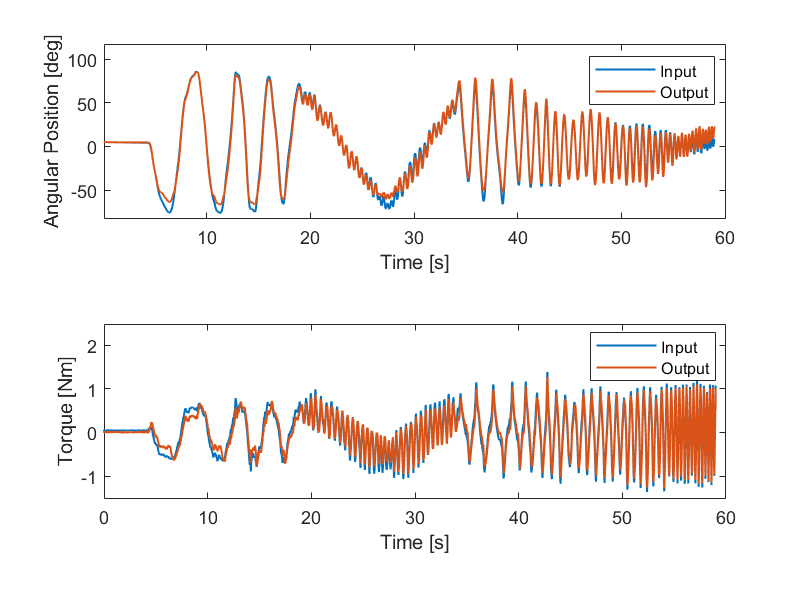}

    \caption{Transmission input/output shaft angular position and torque over time, where input shaft is actuated by hand and output shaft is loaded with an inertia of $0.0387 kg m^{2}$. }
    \label{fig:PosTrqTracking}
    
    \vspace{-2mm}
\end{figure}

To compare the tracking accuracy between the input and output shafts, the transmission was actuated across a large range of motion by hand via a handle on the input shaft, while a load with inertia $0.0387 kg m^{2}$ is attached to the output shaft (Fig. \ref{fig:PosTrqTracking}).
The results show good tracking over the entire motion, with slight tracking error at the position and torque extremities that are likely caused by hysteresis and energy losses.
The close tracking between the input and output torques validate the transmission’s constant mechanical advantage across a large range of motion.

\section{Conclusion}
    In this work, a rolling diaphragm transmission featuring a coaxial opposed rolling diaphragm layout, and a translating core enclosed cable drive, was prototyped and tested.
    The prototype displays low hysteresis, low friction, and good position and torque transparency.
    An automated transmission pressurization and phasing system was also detailed, improving the ease of system setup and maintenance.

    According to the theoretical stiffness model, minuscule amounts of undissolved air can have a significant impact on system stiffness.
    Apart from undissolved air, the rolling diaphragm stiffness contributes the most to system compliance, forming a 'bottom line' on transmission stiffness.
    Further investigation into methods to thoroughly dissolve and bleed air in the transmission, as well as stiffer diaphragm choice, can greatly improve system stiffness. 
    Further understanding the rolling diaphragm in isolation, such as it's damping and rolling friction, may also help identify other design limitations for rolling diaphragm based transmissions.
    
\section*{Acknowledgements}
The authors would like to thank Professor Raymond de Callafon for his advice on system identification, and Alexander Luke for his work on prototype mechanical design.

\clearpage
\balance
\bibliographystyle{ieeetr} 
\bibliography{references}

\end{document}